# Evaluating Performance of Machine Learning Models for Diabetic Sensorimotor Polyneuropathy Severity Classification using Biomechanical Signals during Gait

Fahmida Haque[1], Mamun Bin Ibne Reaz[1]*, Muhammad Enamul Hoque Chowdhury[2]*, Serkan Kiranyaz[2], Mohamed Abdelmoniem[2], Emadeddin Hussein[2], Mohammed Shaat[2], Sawal Hamid Md Ali[1], Ahmad Ashrif A Bakar[1], Geetika Srivastava[3], Mohammad Arif Sobhan Bhuiyan[4], Mohd Hadri Hafiz Mokhtar[1], Edi Kurniawan[5]

[1] Department of Electrical, Electronic and Systems Engineering, Universiti Kebangsaan Malaysia, Bangi 43600, Selangor, Malaysia F.H. (e-mail: fahmida32@yahoo.com) M.B.I.R. (e-mail: mamun@ukm.edu.my), S.H.M.A. (e-mail: sawal@ukm.edu.my), A. A. A B. (e-mail: ashrif@ukm.edu.my), M.H.H.M. (e-mail: hadri@ukm.edu.my)

[2] Department of Electrical Engineering, Qatar University, Doha 2713, Qatar; M.E.H.C (e-mail: mchowdhury@qu.edu.qa), S.K. (e-mail: mkiranyaz@qu.edu.qa), M.A. (e-mail: ma1707821@qu.edu.qa), E.H. (e-mail: eh1707124@qu.edu.qa), M.S. (e-mail: ms1508585@qu.edu.qa).

[3] Department of Physics and Electronics, Dr Ram Manohar Lohia Avadh University, Ayodhya, 224001, India, (e-mail: gsrivastava@rmlau.ac.in).

[4] Department Electrical and Electronic Engineering, Xiamen University Malaysia, Bandar Sunsuria, Sepang 43900, Selangor, Malaysia, (e-mail: arifsobhan.bhuiyan@xmu.edu.my).

5   Research Center for Photonics, National Research and Innovation Agency, Jakarta, 10340, Indonesia, (email: edik004@brin.go.id)

* Corresponding authors: M.B.I.R. (e-mail: mamun@ukm.edu.my), M.E.H.C (e-mail: mchowdhury@qu.edu.qa)

**Abstract**

**Background:** Diabetic sensorimotor polyneuropathy (DSPN) is one of the prevalent forms of neuropathy affected by diabetic patients that involves alterations in biomechanical changes in human gait. In literature, for the last 50 years, researchers are trying to observe the biomechanical changes due to DSPN by studying muscle electromyography (EMG), and ground reaction forces (GRF). However, the literature is contradictory. In such a scenario, we are proposing to use Machine learning techniques to identify DSPN patients by using EMG, and GRF data.

**Method:** We have collected a dataset from the literature which involves, four patient groups: Absent (n=29), Mild (n=18), Moderate (n=16), and Severe (n=14.)  and collected three lower limb muscles EMG (tibialis anterior (TA), vastus lateralis (VL), gastrocnemius medialis (GM) and 3-dimensional GRF components (GRFx, GRFy, and GRFz). Raw EMG and GRF signals were preprocessed, and a newly proposed feature extraction technique scheme from literature was applied to extract the best features from the signals. The extracted feature list was ranked using Relief feature ranking techniques, and highly correlated features were removed. In this study, we have considered different combinations of muscles, and GRF components to find out the best performing feature list for the identification of DSPN Severity. We have trained different ML models to find out the best-performing model and optimized that model. We trained the optimized ML models for different combinations of muscles and GRF components features, and the performance matrix was evaluated.

**Results:** This study has found ensemble classifier model was performing in identifying DSPN Severity, and we optimized it before training. For EMG analysis, we have found the best accuracy of 92.89% using the Top 14 features for features from GL, VL and TA muscles combined. In the GRF analysis, the model showed 94.78% accuracy by using the Top 15 features for the feature combinations extracted from GRFx, GRFy and GRFz signals.

**Conclusion:** The performance of ML-based DSPN severity classification models, improved significantly, indicating their reliability in DSPN severity classification, for biomechanical data.

*Keywords*— Electromyography, biomechanical signal, DSPN, diabetic neuropathy, EMG, biomechanical features, gait. Machine learning, ML, severity grading.

## I. Introduction

Diabetic Sensorimotor Polyneuropathy (DSPN) leads to the progressive loss of somatosensory sensitivity, especially in the lower limbs; this effect may cause functional gait variations and is predominantly related to reductions in joint movement range and active muscle power, and changes in gait mechanics (1). EMG has been widely used by researchers to observe muscle activities and diagnose DSPN in patients (2–5). EMG is an electrophysiological method that measures the electrical activity of muscles. It has been used to evaluate the change in a muscle's electrical activity to diagnose DSPN in clinical research (2–5). Compared with other groups, the DSPN group showed greater stance phase time (6,7) and decreased and delayed lower limb muscle activity; in particular, the vastus lateralis (VL), tibialis anterior (TA), lateral gastrocnemius (LG), gastrocnemius medialis (GM) muscles are the most affected by the progression of neuropathy (2,8,9). Thus, EMG is widely used in different clinical research and trials to diagnose DSPN and observe the biomechanics changes in different muscle activities due to DSPN (2,3,5,9,10). Neuropathies affecting motor nerves will have EMG changes very similar to motor neuron diseases, however, not very popular with patients because of 'their invasive-ness. Due to the changes in biomechanics during gait, a patient suffers from alteration in plantar pressure, kinematic patterns, ground reaction forces, and muscle activities, and long-term alteration in the biomechanics leads to foot ulceration, and in worsening cases amputation of the lower limb (1). Foot ulceration is one of the pervasive types of long-term chronic complication in DN patients, which is an indicator of worsening DN (1,2,11).

The changes in lower limb muscle activities during gait of DSPN patients are also related to other alterations such as higher plantar pressure distribution, greater stance phase, modified ground reaction forces (GRF), and moments of force (7,8,12–16). In a study by Akashi et al. (2), they have shown that DN patients with, and without a history of plantar ulceration had delayed activation peak and a decrease in the second peak of vertical ground reaction force compared with the control group in VL, and GL muscles during gait. Gomes et al. (3) reported delayed activation peaks in the VL, TA, GM, and fibularis longus muscles during gait. Another study by Sacco et al. (7), and Abboud et al. (8) showed delayed muscle activation and decreased muscle amplitude of TA muscle for DSPN patients. Sawacha et al. (5) found an alteration in gait in DSPN patients and suggested that this can be a predictive indicator for the risk of ulceration.

In a meta-analysis, VL, TA, LG, and muscles showed changes in activities due to DSPN. The time of peak muscle activity occurrence was longer for the VL, LG, and GM but reduced for the TA in the DSPN group compared with those in the control group (17). The meta-analysis suggested that compared with that in the DSPN and control groups, the peak occurrence in the TA and GM was non-significantly longer and that for VL in the DSPN group did not differ. Fernando et al. (1) reviewed the biomechanical characteristics of DSPN. Although they considered the EMG dynamics of the three studies, in their meta-analysis, they only observed the TA muscle of the DSPN and control group.

Although these studies are claiming the relationship between the alteration of lower limb muscles, and DSPN, few studies reported prolonged activity in VL, GL, GM, and rectus femoris muscles, and GRF in patients without DSPN (7,16,18,19). With these controversies in studies regarding the biomechanical changes in lower limb muscles for patients with, and without DSPN, such claims are not reliable just by analyzing EMG, and GRF signals. Depending on the variability of the recruited population, for these studies, the biomechanical changes in muscle EMG, and GRF during gait can be varied among the same experimental groups, depending on the patient's ability to adapt to the changes in gait. All the patients do not exhibit any patterned delay in the muscle activation functions which can cause unreliable diagnosis (7,9). With the help of Machine Learning (ML) EMG and GRF signal analysis can be made more robust and overcome the variability in the research regarding the patient's identification.

For the past 50 years, researchers are using EMG signals to identify neuromuscular diseases by collecting the signals from the diagnosed region of interest in the human body and analyzing them manually (20). However, manual analysis of bio-signals like EMG can be challenging. As previously discussed, EMG signals in DSPN patients might not always follow any specific pattern. In such cases, the raw signal analysis does not provide much information from the EMG signals. In such cases, from the EMG, and GRF signals, with the help of machine learning, many important features can be extracted using feature extraction techniques (21).

This research aims to study the performance of ML-based DSPN classifiers using biomechanical features, extracted from EMG and GRF signals during gait. The patients from the collected dataset were classified into different DSPN severity classes based on the MNSI grading system proposed by Watari et al (9). and a newly developed grading system by the authors of the current research (22). As per our knowledge, this is the first ML-based work to classify patients with different degrees of DSPN severity using biomechanical da-ta during gait. After the raw data pre-processing, the feature extraction technique, previously proposed by Johirul et al (23) was used, where two new time-domain features (LMAV, and NSV) with other 17 existing time-domain features were combined. This new feature extraction scheme showed better performance compared with other reported feature extraction schemes in the literature. After feature extraction, the data was processed with the removal of highly correlated features and ranking of the features based on their impact on identifying the different levels of DSPN severity. After finalizing the feature dataset, different conventional machine learning algorithms were trained and optimized to find out the best performing algorithm for the detailed investigation. Different feature combinations extracted from EMG signals from three different lower limb muscles, vastus lateralis (VL), tibialis anterior (TA), and gastrocnemius medialis (GM), and their corresponding 3-dimensional GRF signals (GRFx, GRFy, GRFz) during the gait cycle were investigated to observe the best combination features to identify DSPN severity. Comparative analysis was performed between the developed grading system by the authors of this work (22) and the grading system proposed by Watari et al (9). This ML-based DSPN severity classification technique has the potential to be a secondary decision support system for healthcare professionals in diagnosing different degrees of severity.

II. MATERIALS AND METHODS

In this research, DSPN severity classification using features from EMG and GRF signals during gait with the help of ML has been observed. The biomechanical signals (EMG and GRF) were considered from three lower limb muscles, vastus lateralis (VL), tibialis anterior (TA), and gastrocnemius medialis (GM). The detailed process to achieve that has been discussed below. The overall workflow has been illustrated in Fig. 1.

*A. Dataset Description*

In this research, the dataset was collected from Watari et al study (9). In this study, the authors have collected data from 147 volunteered adults of both genders that were divided into five groups; 30 subjects with no previous history of DM are considered as a control group, and 4 DSPN severity groups were classified: 43 subjects taken from diabetic group and non-neuropathic (in this project classified as absent DSPN), 30 diabetic individuals from mild DSPN, 16 moderate DSPN subjects, and 28 from the severe DSPN subjects. Patients were classified into different DSPN severity groups using a proposed Fuzzy Inference System (FIS) developed by the authors of the study and used the Michigan neuropathy Screening Instrument (MNSI) as the input of their FIS model. From MNSI, they have used, Vibration Perception using a 128Hz tuning fork, Tactile sensitivity using a 10g Semmes-Weinstein monofilament and an MNSI questionnaire. This study performed a de-fuzzification method to transfer the fuzzy system output to a numerical value and produce a 'neuropathy degree score'. The scores (x) are sorted into the disease classes dividing the classes as follows:
1. Absent DSPN: $x \leq 2.5$
2. Mild DSPN: $2.5 < x < 5.0$
3. Moderate DSPN: $5.0 \leq x \leq 8.0$
4. Severe DSPN: $x \geq 8.0$

Patients' severity classes graded were also made available by Watari et al (9) to the authors of the current study and have been considered for comparative analysis in this work. An ML-based DSPN severity grading system has been proposed using MNSI by the authors of the current work and has been reported in another study (22). The proposed grading system was applied, to regrade the patients from Watari et al (9)study, into different DSPN severity classes. Therefore, in this study, two different grading models were used to classify the patients for comparative analysis.

Watari et al. (9) also collected EMG and GRF signals from three lower limb muscles, vastus lateralis (VL), tibialis anterior (TA), and gastrocnemius medialis (GM) during gait for different DSPN severity groups. These three muscles were considered, because of the role of VL and GM in the gait progression, VL and TA represents the knee and ankle im-pact attenuation (9). Table I shows the total number of subjects in each category along with the number of EMG and GRF signal samples per class from this study.

A total of 392 signal samples from 77 patients were collected from Watari et al (9). The authors also provided patients with socio-demographic data along with their MNSI. The dataset also contained the severity grades of the patients, stratified using the Fuzzy based DSPN severity grading system proposed by the authors in their work (9). Three lower limb muscles (GM, TA, VL) EMG signals and their corresponding 3-dimensional GRF signals (GRFx, GRFy, GRFz) during one gait cycle were available in the dataset. signals It is worth mentioning that according to the available data, multiple trials for the same subject are usually recorded.

## B. EMG Signal Processing

In this research, a MATLAB program was devolved to perform the required pre-processing. Samples of raw EMG signals from three lower limb muscles are shown in Fig. 2. Fig. 3 shows the samples of raw signals from 3-dimensional GRF during gait. Firstly, the EMG signals were passed through a built-in de-trending function to remove the undesired low-frequency interference which is the reason behind the non-zero baseline trending in the row signals. This non-zero baseline in EMG signals can lead to miss-interpretations of the EMG signals and their features, which will eventually lead to poor reliability in the diagnoses for clinical applications (24). After detrending, the signals are filtered using a 4th order Butterworth bandpass filter with a default lower cut-off frequency of 4Hz and a higher cut-off frequency of 500Hz. It is worth mentioning, that the lower cut-off frequency can be adjusted later manually using a designated application as will be discussed in a later subsection.

The next step is to apply a notch filter to eliminate the power-line interference noises. In fact, and since the data were collected in the USA where the frequency of the power system is 60Hz, the filter parameters were tuned accordingly to eliminate the fundamental and the related harmonics components. The signal was then rectified and zero-padded. Afterwards, the signal was passed through a high order lowpass Butterworth filter with a cut-off frequency of 5Hz. This method is usually referred to as a low pass filter (LPF) envelope.

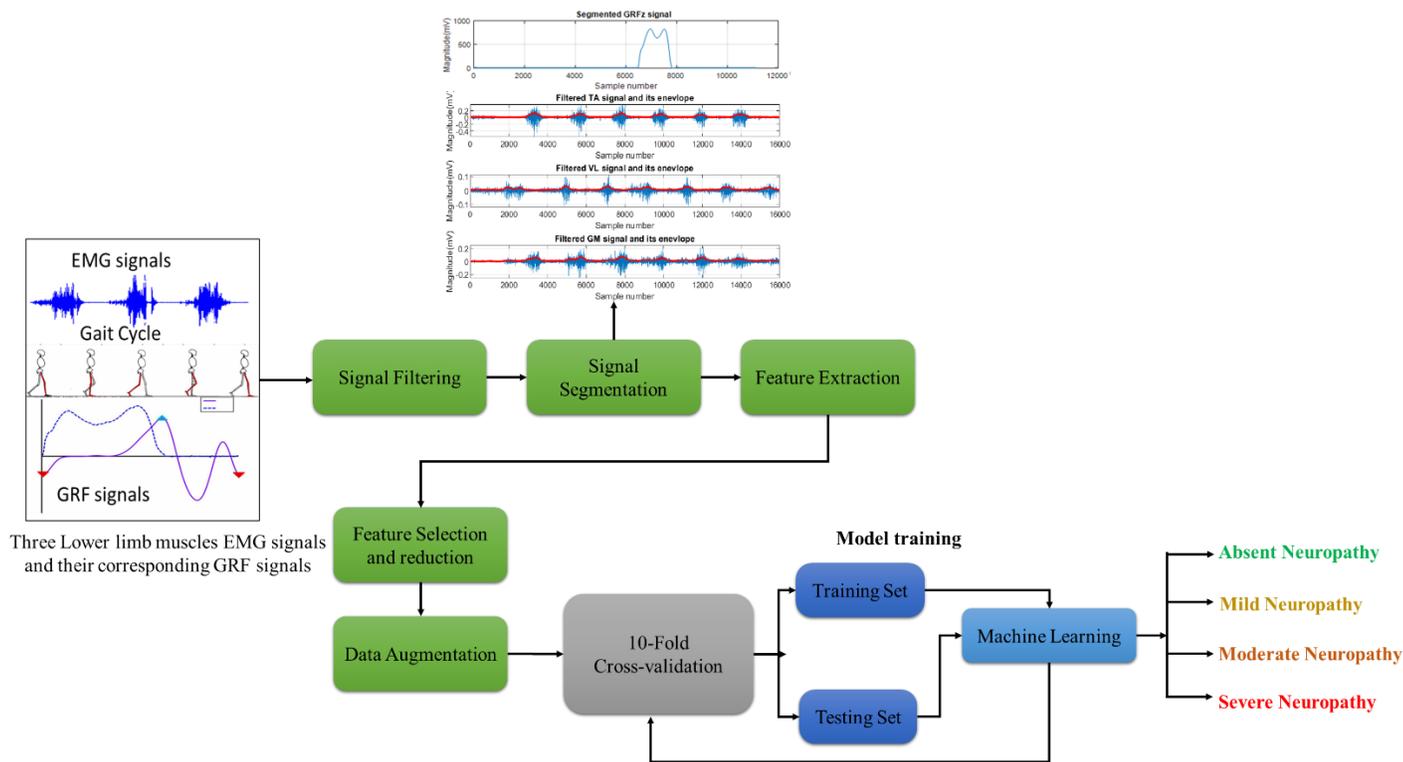

Fig. 1. Block diagram of the overall work process for training and testing different ML models for DSPN severity classification using biomechanical data (EMG and GRF).

## C. Data Segmentation

After the filtering stage, the filtered signals were segmented manually. To facilitate the process of segmenting the signals, a MATLAB program was developed to perform this task. A built-in MATLAB function called "change" was used to segment the GRF signals. This function is designed to detect abrupt changes in the provided signals. The data set contained 3-dimensional GRF signals, mediolateral (GRFx), anteroposterior (GRFy) and vertical (GRFz) GRF and these signals were available for only one stance (i.e., from one gait cycle) was recorded. Using the built-in MATLAB function called change, the GRFz signal for one stance phase was segmented for each sample. The detection technique is selected to be linear, to find abrupt changes in the slope and the intercept of the signals rather than the mean of the data. Moreover, a threshold of 200 samples was selected to limit the number of detected change points in the signal. The segmented GRFz signal was used to segment the other two GRF components GRFx, and GRFy.

Envelope signals were generated for all the EMG signals for manually checking the signals before segmentation as illustrated in Fig. 4. A manual check was then performed to ensure reliable segmentation of EMG signals. This manual check was performed by visual inspection of the signals using a special MATLAB function to create a customizable polyline that was used to interactively modify the initial indices in the signals. Each EMG signal was segmented considering the start of one active phase or burst till the start of another active phase. So, each segment consists of one bust and one silence phase. Typically, four segments were expected from each muscle signal. By observing the envelope of the signals, the activation of GM is expected to be synchronized with the starting of the z-component in the GRF signals during the "stand phase". Then, the activation of the TA muscle is usually before the VL and GM, and the activation of the VL is started before the GM. Afterwards, signal data were visualized to understand the overall characteristics and behaviour of the signals. This will help in debugging and boosting the performance of the ML models. To accomplish this, a MATLAB script was developed to calculate and plot the mean and the stander deviation of the signal in each class. Initially, envelope signal was detected for all severity classes EMG signals. Then the signals were smoothened and normalized mean using 'rms'.

TABLE I
DATASET DESCRIPTION COLLECTED FROM WATARI ET AL. (9)

| DSPN severity | No patients per class | No EMG and GRF samples per class |
|---|---|---|
| Absent | 29 | 142 |
| Mild | 18 | 93 |
| Moderate | 16 | 85 |
| Severe | 14 | 72 |
| Total | 77 | 392 |

## D. Feature Extraction

In this stage, a set of descriptive and informative features were extracted from the filtered and segmented signals, since these features are expected to facilitate the classification process especially when classical ML algorithms are utilized. That was because having the raw EMG signal as an input for pattern recognition and classification is not efficient. Thus, applying feature extraction is essential to map the row signals into lower dimension feature vectors (25). One of the most important and challenging steps in EMG processing is feature extraction since the target is to transform the raw signals into a reduced representative set of features.

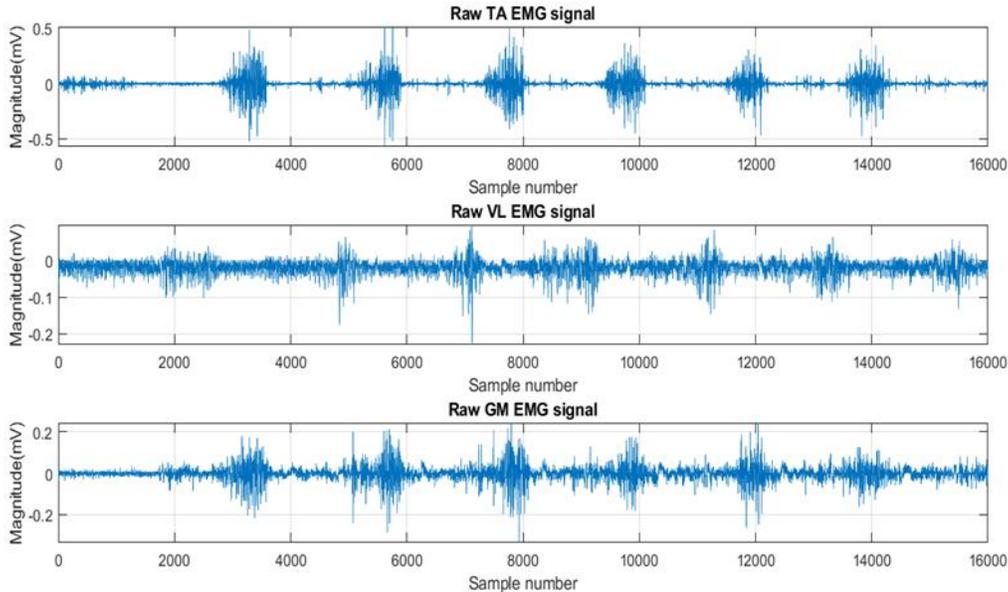

Fig. 2. Sample EMG signals from TA (top), VL (middle) and GM (bottom) muscles.

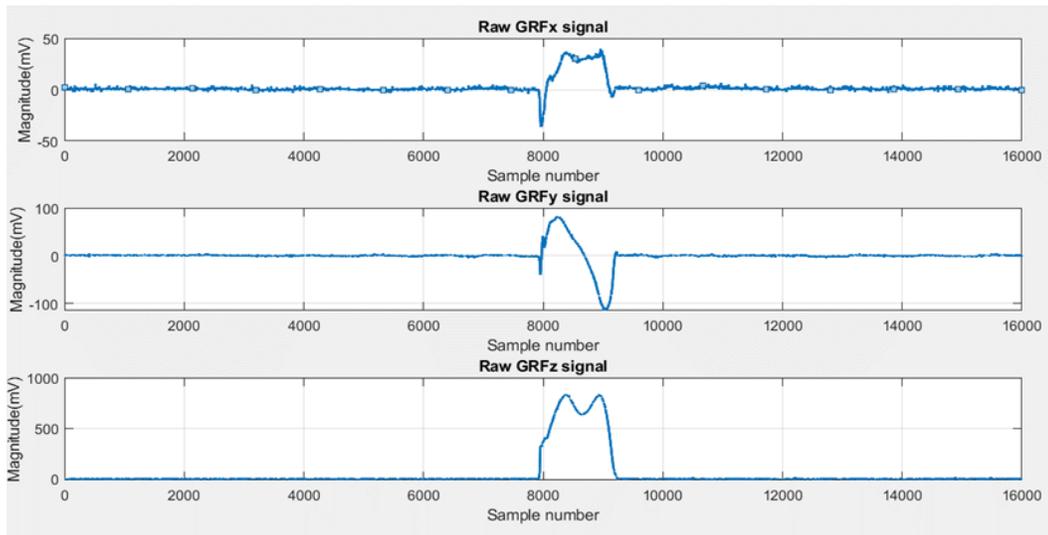
Fig. 3. Sample GRF signals from GRFx (top), GRFy(middle) and GRFz (bottom) muscles.

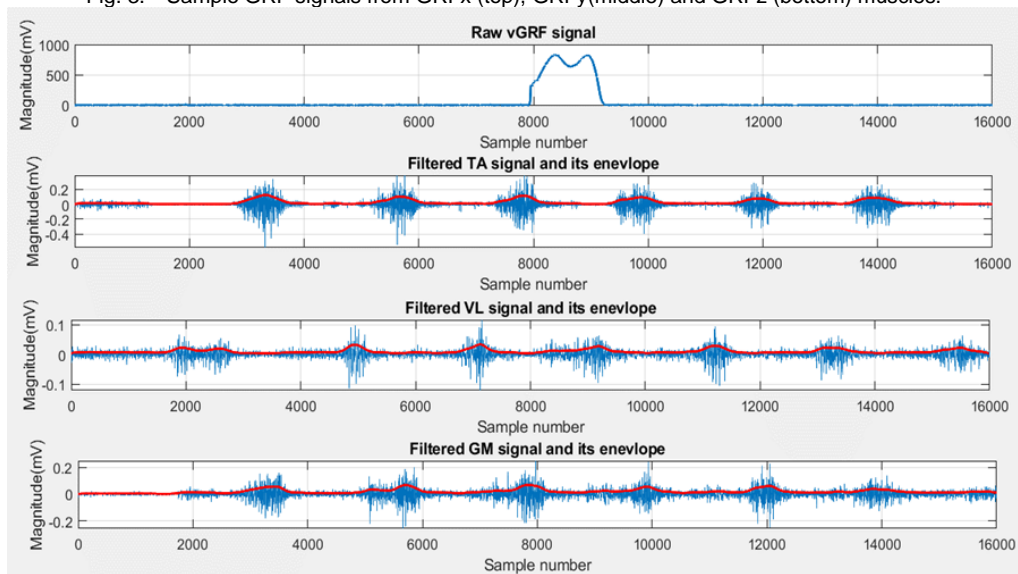
Fig. 4. Sample of filtered EMG signals with their envelopes and corresponding GRFz signals.

Pre-processed EMG and GRF signals were used for feature extraction. In a previous study, Johirul et al (23) proposed two new time-domain features, the log of the mean absolute value (LMAV), and nonlinear scaled value (NSV) along with 17 commonly used time-domain feature extraction schemes. The details regarding these two features can be found in (23). In their work, they showed that these time-domain features incorporated with other 17-time domain (TD) feature from existing literature exhibited better performance in the pattern recognition problem. So, we are introducing these new proposed features in the disease diagnosis domain, with other existing features in identifying DSPN severity. In this research, we have used nineteen-time domain features previously reported by Johirul et al (23). The extracted time domain features are the log of the mean absolute value (LMAV), nonlinear scaled value (NSV), waveform length (WL), Wilson amplitude (WAMP), slope sign changes (SSC), number of zero crossings (ZC), mobility (MOB), complexity (COM), and skewness (SKW), and four autoregressive coefficients (AR1, AR2, AR3, and AR4), Moment (0th,2nd, 4th, and 6th order) (m0-m6), Amplitude Change (1, and 2) (AC1, AC2). All of the features were extracted using in-house-written MATLAB code.

### E. Feature Selection

The main target in the feature selection stage is to eliminate all the redundant and highly correlated features and reduce the extracted features space into a smaller subset that is highly distributive and informative to ensure high classification performance (26). In other words, it's a method to reduce the number of input variables to a level where it is believed that they are useful to predict the targeted variables. Feature selection methods are applied to remove irrelevant features to end up with a better performing model, an easier understanding model, and a model that runs faster (26). In some cases, the performed feature input holds a huge number of variables or features that may slow the process of development and training of the model, resulting in a large amount of system memory. Another reason feature selection prevents, degrading the model is by having input variables that are not relevant to the targeted variables or having overfitting.

To discover the best techniques careful systematic experimentation was done.

In this research, we have used the Relieff technique which is a filter type feature selection method. This technique uses pairwise distance to find the distance-based supervised models between observations to predict response and estimate the feature importance (26). The algorithm computes instances at r, random, computes their nearest neighbour, and varies a feature weighting vector to give more weight to features that distinguish the instance from the neighbour of different classes.

The correlations among different features were observed for both EMG and GRF signals. Highly correlated feature elimination is performed before feature ranking to not impact the feature ranking algorithm. The correlation matrix between features is calculated using pairwise linear correlation. A threshold of 0.9 is considered a highly collated feature, and one of them is dropped from the feature vector.

After removing highly correlated features, datasets were prepared for three individual lower limb muscles (GM, VL, and TA), and three-dimensional GRF (GRFx, GRFy, and GRFz) for the remaining features. Using the Relieff feature selection technique, the remaining features from EMG and GRF were ranked based on their importance index in DSPN severity identification.

### F. ML Model Development Using Biomedical Data

After removing highly correlated features and ranking the remaining features a wide range of datasets were prepared by combining different muscle features for EMG and different component GRF features. It can be considered as channels based on the number of muscle features or GRF components features were considered. The ML models were also trained on the different channel datasets for both EMG and GRF and their performance was analyzed to find out the best performing model. A detail of the dataset and its corresponding features are listed in Table II for EMG and Table III for EMG.

TABLE II
NUMBER OF FEATURES EXTRACTED FROM LOWER LIMB MUSCLE EMG FOR A DIFFERENT COMBINATION

| Combination | Muscles | No. of Features | No. of Significant Features |
|---|---|---|---|
| One Channel | GL | 19 | 10 |
| | VL | | 7 |
| | TA | | 10 |
| Two-Channel | GL-VL | 38 | 15 |
| | GL-TA | | 17 |
| | TA-VL | | 17 |
| Three Channel | GL-VL-Ta | 57 | 22 |

For the classification of DSPN severity classes (absent, mild, moderate, severe) machine learning algorithms were used. To find out the best performing algorithm, six different algorithms: Discriminant analysis classifier (DAC), Ensemble classification model (ECM), k-nearest neighbour model (KNN), Naive Bayes classifier (NBC), Support vector machine classifier (SVM), and Random Forest (RF) were trained using the EMG, and GRF features. Fitcauto function from MATLAB 2020b was used for training and hyperparameter tuning of the models. We have optimized all the algorithms by using Bayesian optimization.

For each learner, the function was optimizing all possible hyperparameters by setting the hyperparameter option to all, using the Bayesian optimization. Tuning of the hyperparameters of the ML algorithm was implemented in MATLAB version 2020b (MathWorks, USA). After finding the best performing model, it was used to classify the patients into four classes, using the EMG, and GRF features data. The best performing ML algorithm was trained in three different ways. The detailed training process is described below:

*Single Channel:* Initially, the ML model was trained for individual muscle or GRF component data. So, the ML model was trained for three individual muscle features (GM, VL, and TA), and three GRF component features (GRFx, GRFy, and GRFz) separately, and observed the classification performance.

*Two Channel:* Secondly, the ML model was trained in a combination of two muscles (GM, and VL, GM, and TA, and TA, and VL) or two GRF components (GRFx, and GRFy, GRFx, and GRFz, and GRFy, and GRFz).

*Three Channel:* Lastly, the performance of the ML model was observed with all three muscles (GL, VL, and TA) or three GRF (GRFx, GRFy, and GRFz) components combined.

TABLE III
NUMBER OF FEATURES EXTRACTED FROM LOWER LIMB MUSCLE GRF FOR A DIFFERENT COMBINATION

| Combination | GRF components | No. of Features | No. of Significant Features |
|---|---|---|---|
| One Channel | GRFx | 19 | 14 |
| | GRFy | | 14 |
| | GRFz | | 14 |
| Two-Channel | GRFx-GRFy | 38 | 21 |
| | GRFx-GRFz | | 19 |
| | GRFy-GRFz | | 19 |
| Three Channel | GRFx-GRFy-GRFz | 57 | 25 |

The ML model was trained and validated with 10-fold stratified cross-validation. As the dataset was imbalanced, the dataset was augmented using Synthetic Minority Oversampling Technique (SMOTE) technique balance dataset (27). The performance of the ML algorithms was evaluated using different evaluation parameters such as accuracy, sensitivity, specificity, precision, and F-1 score. The model performance in classifying three experimental groups was analyzed using the incremental feature search technique for all the ranked features in an ascending manner, starting from only the top 1 feature, and then going on from the top 2 features and going up. Here, we wanted to investigate, the number of top-ranked features combined to obtain the best performance for respective cases. All processes were conducted for EMG, and GRF data separately. The performance of the ML model was compared for the proposed grading model (22) and. the grading model by Watari et al. (9) for the comparative analysis between two grading systems.

## III. Results

### A. Patients Characteristics

The sociodemographic and clinical characteristics of the patients from the Watari et al (9) study have been discussed and also illustrated in Table IV. All the recruited patients in this study have Type 2 diabetes mellitus (DM). So significant difference was present among the 4 DSPN severity groups in age, mass, height, and body mass index (BMI). Severe group patients showed slightly higher BMI compared with other groups. Also, it is evident that, with the longer duration of DM, the chance of increasing the severity of the DSPN. Also, with the progression of severity from absent to severe, the level of glycemia in patients is higher.

For performance analysis of ML models for DSPN severity classification using biomechanical data, three lower limb muscles (GM, VL, TA) and 3-dimensional GRF (GRFx, GRFy, GRFz) signals were used to extract the biomechanical features during gait. The patients were graded into different severity classes using the Watari et al proposed grading models (9). The performance analysis of the ML modes for both EMG and GRF features are being discussed separately in the below sections.

TABLE IV
SOCIODEMOGRAPHIC AND CLINICAL DATA OF EXPERIMENTAL GROUPS FROM WATARI ET AL. (9) STUDY

| | | DSPN Severity classes | | | | $p^a$ |
|---|---|---|---|---|---|---|
| | | Absent (n=29) | Mild (n=18) | Moderate (N=16) | Sever (n=14) | |
| Gender | Female/Male | 14/15 | 8/10 | 10/6 | 2/12 | |
| Age (Years) | Mean ± SD | 57.07±5.93 | 54.83±8.03 | 58.19±5.33 | 57.29±4.76 | 0.65 |
| | Min, Max | 43, 64 | 31, 64 | 46, 66 | 49, 64 | |
| Mass (Kg) | Mean ± SD | 75.50±14.47 | 78.612±17.50 | 74.27±10.98 | 82.11±13.08 | 0.33 |
| | Min, Max | 46.4, 104.6 | 49.9, 124.9 | 59.2, 92.5 | 63.2, 101.5 | |
| Hight (m) | Mean ± SD | 1.629±0.10 | 1.65±0.10 | 1.62±0.08 | 1.67±0.08 | 0.38 |
| | Min, Max | 1.425, 1.825 | 1.455, 1.89 | 1.48, 1.82 | 1.545, 1.89 | |
| BMI (Kg/m$^2$) | Mean ± SD | 28.31±3.9 | 28.78±4.69 | 28.49±4.19 | 29.36±3.58 | 0.50 |
| | Min, Max | 20.62, 39.86 | 22.74, 39.87 | 21.93, 36.59 | 23.25,35. 12 | |
| DM duration (Years) | Mean ± SD | 6.45±6.26 | 8.28±5.54 | 13.06±9.12 | 15.57±8.74 | <0.001 |
| | Min, Max | 0.08, 26 | 1, 23 | 2, 34 | 2, 35 | |
| Glycemia (mg/dL) | Mean ± SD | 131.±47.3 | 141.7± 67.9 | 174.9±75.9 | 242.4±102.8 | <0.001 |
| | Min, Max | 66, 320 | 47, 375 | 72, 377 | 118, 390 | |

[a] Pearson Chi-Square test

### B. Analysis of EMG Signals Corresponds to DSPN Severity

In Fig. 5 the mean of the EMG signals (in each class) is plotted for all the available muscles. From these plots, some important characteristics regarding the signals can be observed, such as the relative reduction in the peak magnitude of the TA signals in DSPN subjects compared to non-DSPN subjects. Moreover, for TA, the delay in peak activation continuous to be a distinguishable feature in diabetic subjects, and in the case of TA this is clear when considering the onset peak activation as compared to the absent group A relative delay in the peak activation of the GM signals between the DSPN classes (mild, moderate and severe) can also be considered as a sign to differentiate between classes, however, absent class exhibited a delayed peak activation in comparison with other three severity classes. Lastly, in VL signals, moderate and severe subjects are presenting the expected delay pattern in the pack activation with a relative reduction in the amplitude too. These observations are expected since an alteration in the shock absorbance mechanism at the ankle countered by a higher dependency on the knee to support the body weight during the weight acceptance phase of the gait is widely common during the early stage of DM (9). Additionally, a delayed peak and a decrease in the magnitude of the activation in GL signals are associated with severe DSPN subjects.

This delay in the muscle activation patterns especially in DSPN subjects is expected since an inefficiency in gait motion is associated with more advanced DSPN cases, such alterations in the locomotor pattern could be due to alteration in the shock attenuation, and the mechanical overloading during the heel strike, or even some changes in ankle ex-tensor activities during gait (3), (9).

### C. ML Model Selection

Six different algorithms: Discriminant analysis classifier (DAC), Ensemble classification model (ECM), k-nearest neighbour model (KNN), Naive Bayes classifier (NBC), Support vector machine classifier (SVM), and Random Forest (RF) were trained and optimized for the EMG, and GRF extracted features to find out the best performing model for the detailed analysis. For EMG signals, the ECM algorithm outperformed other algorithms using EMG, and GRF data. The performance of these six classifiers for features extracted from EMG and GRF signals has been listed in Supplementary Tables S1 and S2. The ECM algorithm hyperparameters were then optimized for both EMG and GRF features. For EMG features, the optimized ECM model used the 'AdaBoostM2' as 'method', 'NumLearningCycles', set to 305, 'LearnRate', set to 0.96, 'MaxNumSplits', set to 71, 'NumVariablesToSample', set to 'all'. AdaBoostM2 uses an Adaptive boosting technique for multiclass classification problems. For GRF features, the optimized ECM model used the 'Bag' as 'method', 'NumLearningCycles', set to 305, 'LearnRate', set to 0.96, 'MaxNumSplits', set to 430, 'NumVariablesToSample', set to 5. In the "Bag" method, predictors are randomly selected at each split using a random forest algorithm. Both optimized ECM algorithms were trained for using different feature combinations extracted and processed from the EMG, and GRF signals, respectively. The below section will be highlighting the best performing results by the EMG, and GRF feature sets.

### D. Performance Analysis of ML-Based DSPN Severity Classification Using EMG Features

*1) Fuzzy based DSPN severity grading system* (9)

The EMG dataset initially consists of 19 TD features for each muscle. Here we have implemented the study channel-wise, assuming each signal features a channel data. So, in one channel analysis, we have considered three lower limb muscles, GL, VL, and TA with 19 features proposed by Johirul et al. (23) individually. In two-channel analysis, we have taken a combination of two muscle features at a time, and for three-channel, we have considered all three muscle features. EMG feature dataset was ranked using the Relieff feature ranking technique based on their importance in identifying DSPN severity after removing highly correlated features. The number of features before and after feature ranking and removal of highly correlated features are listed in Table II. The feature list after ranking and highly correlated feature removal can be found in Supplementary Table S3-S5. The results from Relieff feature ranking techniques for features from GM, VL and TA muscles and also for combined muscles features from GM and VL; GM and TA; VL and TA; and GM, TA and VL muscles were plotted to find out the importance of the features for identifying DSPN. Here, the result from feature ranking for the best performing features combination (feature combinations from GA, TA and VL muscles EMG) is shown in Fig. 6.

The dataset consists of top-ranked features after removing highly correlated features, which were used to train the optimized ML model, and their performance was analyzed. ROC curve, as well as other evaluation matrices, were generated using top feature combinations starting from top 1 features, and so on incrementally, to observe the performance of the model. In Table V we have summarized the best performance by the ECM classifier model in identifying DSPN severity patients using EMG feature combinations from three different lower limb muscles.

From Table V, it can be observed that, in one channel analysis, the ML model achieved, 77.64% accuracy by Top 10 features combination from GM muscle, 69.18% accuracy for Top 8 features combination from TA muscles, 77.64% accuracy by Top 10 features combination from VL muscle. For two-channel analysis, the ML model achieved 79.75% accuracy by Top 14 features combination from GM and VL muscles, 80.45% accuracy for Top 15 features combination from GM and TA muscles, 76.06% accuracy by Top 10 features combination from VL and GM muscles. From the three-channel analysis, the ML model achieved 82.56% accuracy by Top 13 features combined from GM, TA and VL muscles. So, it can be visible from this analysis that all three muscle features combined are provided with the best performance of 82.56% accuracy. The ROC curve has been shown for the best forming model which in this case, is from the three-channel analysis, in Fig 7. The ROC and AUC are calculated for the correct identification of non-DSPN patients by the ML models.

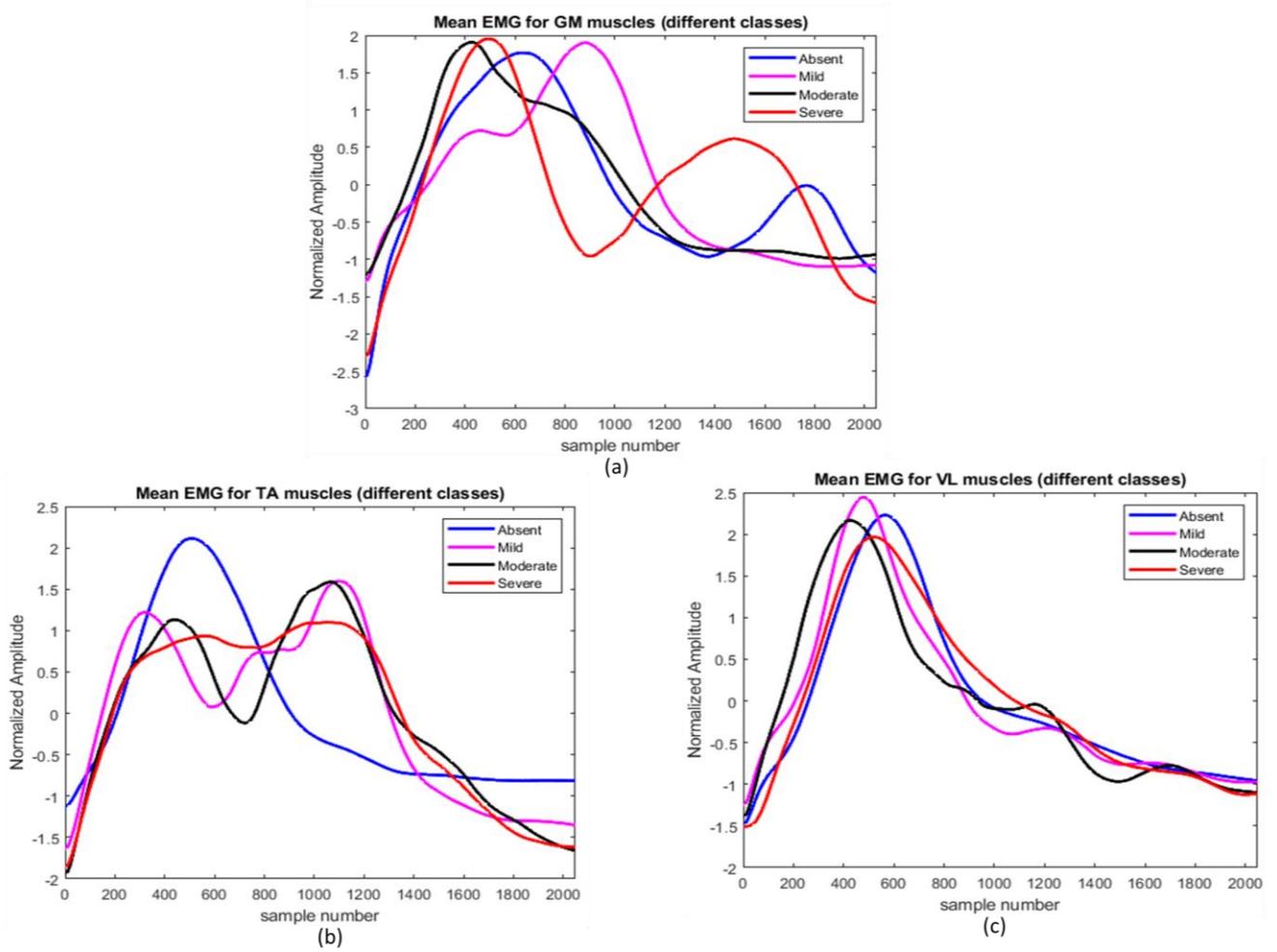

Fig. 5. Smoothened & Normalized mean and stander deviation of the EMG signals for (a) GM (b) TA, and (c) VL muscles for all DSPN severity classes.

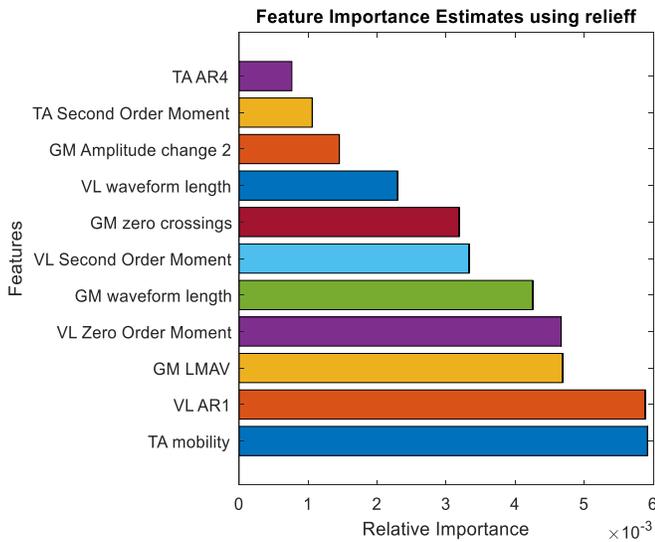

Fig. 6. Top-ranked features extracted from GM, TA, and VL muscles EMG signal by Relieff feature selection techniques for the proposed grading system by Watari et al. (9)

*2) Proposed DSPN Severity grading system* (22)

The impact of the newly proposed grading system on the performance of the ML-based DSPN severity classification (22) using biomechanical data has been studied. Patients from Watari et al (9) study are being reclassified using the developed ML-based DSPN severity grading system using MNSI. Now from the pre-processed EMG and GRF signals, features were extracted, and highly correlated features were excluded. After that, features were ranked using the Relieff technique and the dataset was prepared for the machine learning model training. Here, the result from feature ranking for the best performing features combination (feature combinations from GA, TA and VL muscles EMG) is shown in Fig. 8. The feature list after ranking and highly correlated feature removal can be found in Supplementary Table S6-S8.

It can be visible in Fig. 6 and Fig. 8; that the list of top-ranked features changes due to the changes in patients' severity grades. The dataset consists of top-ranked features after removing highly correlated features, which were used to train the optimized ML model, and their performance was analyzed. ROC curve, as well as other evaluation matrices, were generated using top feature combinations starting from Top 1 features, and so on incrementally, to observe the performance of the model. In Table VI we have summarized the best performance by the ECM classifier model in identifying DSPN severity patients using EMG feature combinations from three different lower limb muscles.

From Table VI, it can be observed that, in one channel analysis, the ML model achieved, 86.89% accuracy by Top 10 features combination from GM muscle, and similar accuracy was achieved by TA and VL also. For two-channel analysis, the ML model achieved 79.75% accuracy by Top 13 features combination from GM and VL muscles, 91% accuracy for Top 13 features combination from GM and TA muscles, 88.11% accuracy by Top 14 features combination from VL and GM muscles. From the three-channel analysis, the ML model achieved 92.89% accuracy by Top 14 features combined from GM, TA and VL muscles.

So, it can be visible from this analysis that all three muscle features combined are provided with the best performance of 92.89% accuracy. The ROC curve has been shown for the best performing model which in this case, is from the three-channel analysis, in Fig. 9. The ROC and AUC are calculated for the correct identification of non-DSPN patients by the ML models.

TABLE V
PERFORMANCE ANALYSIS OF ML MODEL USING DIFFERENT COMBINATIONS OF EMG FEATURES FROM THREE LOWER LIMB MUSCLES FOR THE PROPOSED GRADING SYSTEM BY WATARI ET AL. (9).

| Channel | Muscles considered | Features | Accuracy (%) | Sensitivity (%) | Specificity (%) | Precision (%) | F1-Score (%) | AUC |
|---|---|---|---|---|---|---|---|---|
| One Channel | GM | Top 10 | 77.64±2.81 | 59.90±6.56 | 92.49±1.33 | 77.69±2.83 | 77.77±2.87 | 0.90 |
| | TA | Top 8 | 69.18±6.04 | 57.73±13.13 | 86.62±3.28 | 69.22±6.06 | 69.24±6.17 | 0.83 |
| | VL | Top 10 | 77.64±2.81 | 59.90±6.56 | 92.49±1.33 | 77.69±2.83 | 77.76±2.86 | 0.91 |
| Two Channel | GM, VL | Top 14 | 79.75±5.32 | 66.92±2.83 | 92.01±4.52 | 79.74±5.29 | 79.89±5.26 | 0.91 |
| | GM, TA | Top 15 | 80.45±4.15 | 69.75±9.13 | 94.13±2.77 | 80.51±4.16 | 80.55±4.12 | 0.93 |
| | VL, TA | Top 10 | 76.06±2.38 | 60±14.69 | 90.39±5.85 | 76.13±2.31 | 76.49±2.38 | 0.88 |
| Three Channel | GM, VL, TA | Top 13 | 82.56±4.38 | 71.87±4.64 | 92.02±2.55 | 82.56±4.43 | 82.58±4.40 | 0.92 |

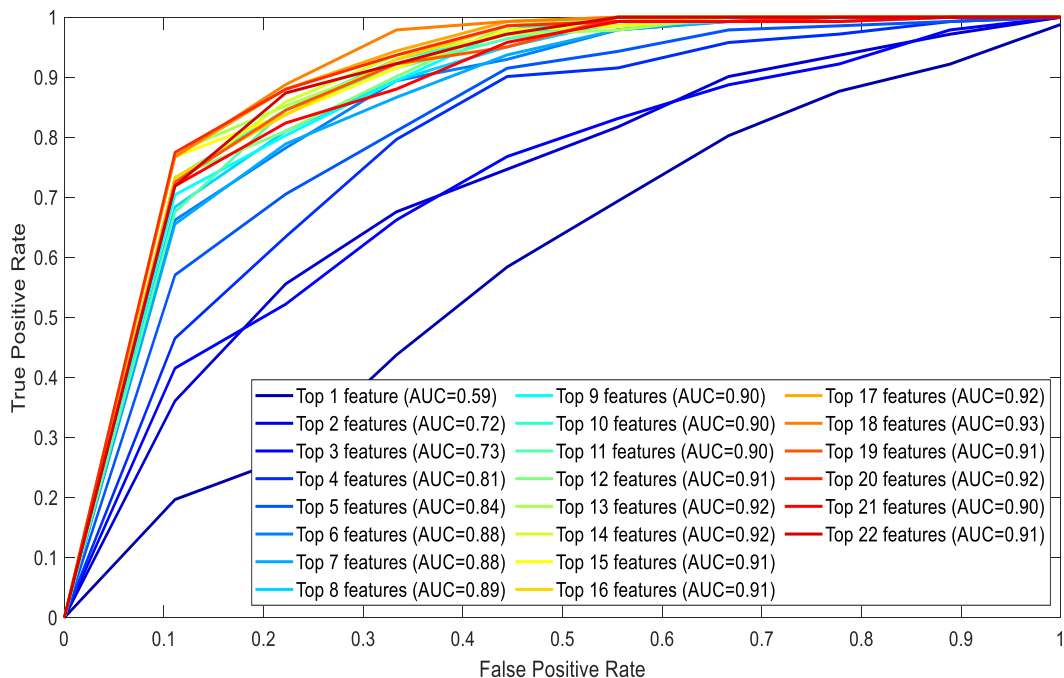

Fig. 7. ROC curve for top-ranked feature combinations from the GM, VL and TA muscles EMG signal for the proposed grading system by Watari et al. (9)

### E. Performance Analysis of ML-Based DSPN Severity Classification Using GRF features

*1) Fuzzy based DSPN severity grading system* (9)

Similar to the EMG analysis, the GRF dataset initially consists of 19 TD features for each GRF component signal (GRFx, GRFy, GRFz). The study is conducted channel-wise, assuming each signal features channel data. So, in one channel analysis, we have considered three individual GRF components separately, GRFx, GRFy, and GRFz with 19 features proposed by Johirul et al. (23). In two-channel analysis, we have taken a combination of two GRF components features at a time, such as GRFx and GRFy; GRFx and GRFz; GRFy and GRFz and for three-channel, we have considered all three GRF components (GRFx, GRFy and GRFz) features combined. GRF features from different channels were ranked using the Relieff feature ranking technique based on their importance in identifying DSPN severity after removing highly correlated features. The number of features before and after feature ranking and removal of highly correlated features has been discussed in the previous section. The results from Relieff feature ranking techniques for features from one, two and three channels were plotted to find out the importance of the features for identifying DSPN. Here, the result from feature ranking for the best performing features combination (feature combinations from GRFx, GRFz and GRFy components), is shown in Fig. 10. The feature list after ranking and highly correlated feature removal for all channels can be found in Supplementary Table S9-S11.

The dataset consists of top-ranked features after removing highly correlated features, which were used to train the optimized ML model, and their performance was analyzed. ROC curve, as well as other evaluation matrices, were generated using top feature combinations starting from top 1 features, and so on incrementally, to observe the performance of the model. In Table VII, we have summarized the best performance by the ECM classifier model in identifying DSPN severity patients using different GRF feature combinations from 3 GRF components signals in different combinations (channel-wise).

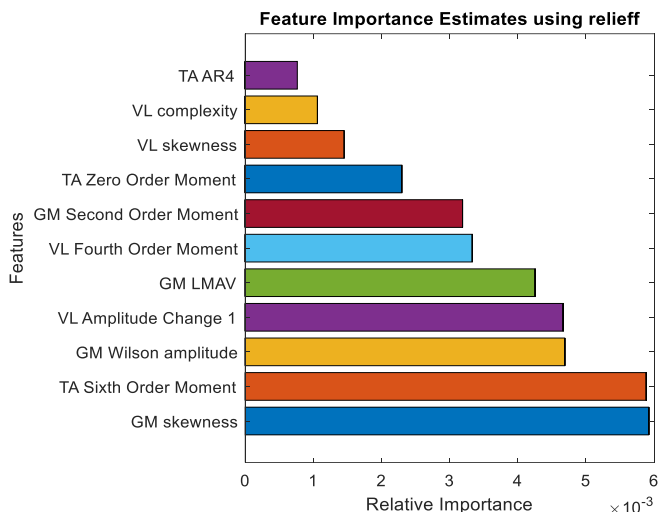

Fig. 8. Top-ranked features extracted from GM, TA, and VL muscles EMG signal by Relieff feature selection techniques for the proposed grading system (22).

For two-channel analysis, the ML model achieved 85.39% accuracy by Top 15 features combination from GRFx and GRFz signals, 86.62% accuracy for Top 15 features combination from GRFx and GRFy signals, 85.21% accuracy by Top 15 features combination from GRFy and GRFz signals. From the three-channel analysis, the ML model achieved 86.62% accuracy by Top 18 features combination from GRFx, GRFy and GRFz. Now, it is visible that, the two models have similar accuracy, for feature combinations from GRFx and GRFy: and feature combinations from GRFx, GRFy and GRFz. However, the model developed using feature combinations from GRFx and GRFy requires a smaller number of features (Top 15) in comparison with three-channel (Top 18). So, in this case, feature combinations from GRFx and GRFy can be selected as the best performing model. The ROC curve has been shown for the best forming model which in this case, is from the two-channel analysis (GRFx and GRFy), in Fig. 11. The ROC and AUC are calculated for the correct identification of non-DSPN patients by the ML models.

*2) Proposed DSPN Severity grading system (22)*

Similar procedures have been followed for the GRF analysis for the developed grading system (22) as is done for the grading system by Watari et al (9). The results from Relieff feature ranking techniques for features from one, two and three channels were plotted to find out the importance of the features for identifying DSPN. Here, the result from feature ranking for the best performing features combination (feature combinations from GRFx, GRFz and GRFy components), is shown in Fig. 12. The feature list after ranking and highly correlated feature removal for all channels can be found in Supplementary Table S12-S14.

In Table VIII we have summarized the best performance by the ECM classifier model in identifying DSPN severity patients using different GRF feature combinations from 3 GRF components signals in different combinations (channel-wise).

TABLE VI
PERFORMANCE ANALYSIS OF ML MODEL USING DIFFERENT COMBINATIONS OF EMG FEATURES FROM THREE LOWER LIMB MUSCLES FOR PROPOSED GRADING SYSTEM (22)

| Channel | Muscles considered | Features | Accuracy (%) | Sensitivity (%) | Specificity (%) | Precision (%) | F1-Score (%) | AUC |
|---|---|---|---|---|---|---|---|---|
| One Channel | GM | Top 10 | 86.89±2.14 | 86.89±2.14 | 95.56±1.89 | 86.89±2.13 | 86.95±2.12 | 0.94 |
| | TA | Top 10 | 86.89±1.60 | 86.89±1.60 | 95.11±1.44 | 86.89±1.60 | 86.90±1.65 | 0.94 |
| | VL | Top 10 | 86.89±2.14 | 86.89±2.14 | 95.56±1.89 | 86.89±2.13 | 86.9±2.12 | 0.94 |
| Two Channel | GM, VL | Top 13 | 91.67±3.04 | 91.67±3.04 | 98.07±1.12 | 91.67±3.04 | 91.75±3.03 | 0.98 |
| | GM, TA | Top 13 | 91±1.38 | 91±1.38 | 96.89±1.92 | 91±1.33 | 91.15±1.44 | 0.97 |
| | VL, TA | Top 14 | 88.11±3.37 | 88.11±3.37 | 94.96±1.5 | 88.11±3.37 | 88.20±3.35 | 0.96 |
| Three Channel | GM, VL, TA | Top 14 | 92.89±2.76 | 92.89±2.76 | 98.07±1.62 | 92.89±2.76 | 92.92±2.78 | 0.98 |

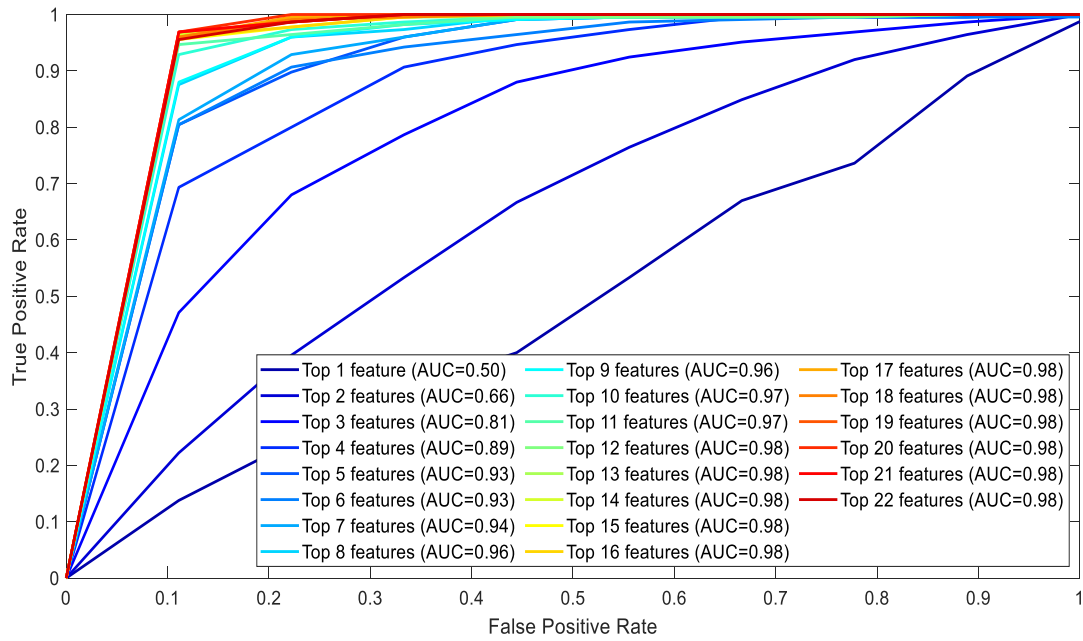

Fig. 9. ROC curve for top-ranked feature combinations from the GM, VL and TA muscles EMG signal for the proposed grading system (22).

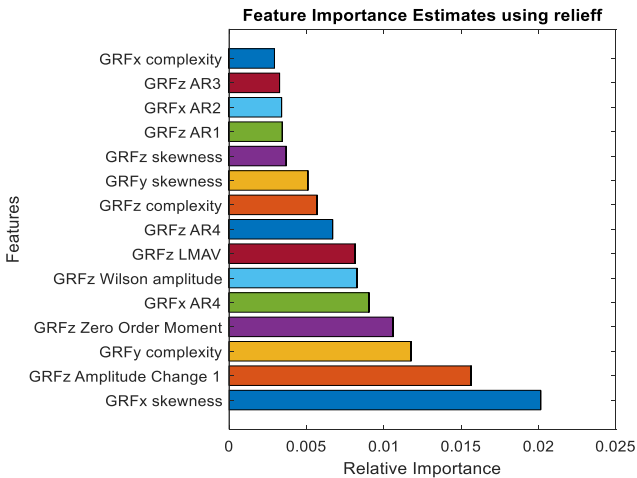

Fig. 10. Top-ranked features extracted combined from GRFx, GRFy, and GRFz components signal by Relieff feature selection techniques for the proposed grading system by Watari et al. (9)

From Table VI, it can be observed that, in one channel analysis, the ML model achieved, 90.11% accuracy by Top 13 features combination from GRFx, 89.11% accuracy for Top 10 features combination from GRFy, 86.44% accuracy by Top 10 features combination from GRFz. For two-channel analysis, the ML model achieved 94.78% accuracy by Top 15 features combined from GRFx and GRFz signals, 92.56% accuracy for Top 16 features combined from GRFx and GRFy signals, 93.78% accuracy by Top 13 features combined from GRFy and GRFz signals. From the three-channel analysis, the ML model achieved 94.67% accuracy by Top 15 features combination from GRFx, GRFy and GRFz. So, in this case, feature combinations from three GRF components outperformed the other models. The ROC curve has been shown for the best forming model which in this case, is from the three-channel analysis (GRFx, GRFy, GRFz), in Fig. 13.

### 3) Comparative analysis between Proposed (22) and Fuzzy (9) based grading system

The performance of ML-based DSPN severity classification using biomechanical feature data has been studied for two different grading techniques. Now for easy observation, the performance of ML models using biomechanical features is being studied for both grading techniques, and comparative analysis is performed to find the better grading system, which can be applied for DSPN severity stratification using MNSI. Fig. 14 illustrate the comparative analysis of the ML model's performance for two grading system, based on the classifier's performance accuracy using different EMG feature combinations from three different lower limb muscles during gait. From Fig. 14, it is visible that, for all the muscle feature combinations, the ECM classifier with patients graded by the proposed MNSI grading system, achieved higher accuracy in comparison to the model with patients classified by Watari et al (9) grading system. The highest accuracy (92.89%) was achieved by the ECM classifier trained with EMG features extracted from all three muscles combined (GM-TA-VL) for the proposed grading model. Also, it is observed that the proposed grading model outperformed the Watari (9) grading system for ML-based DSPN severity classifications for all possible combinations of EMG features during gait.

The comparative analysis was also done for the GRF study, Fig. 15 illustrate the comparative analysis among ML models' performance for two grading system, based on the classifier's performance accuracy using different GRF feature combinations from three-dimensional GRF (GRFx, GRFy GRFz) components, during gait. From Fig. 15 it is visible that, the ECM classifier with patients graded by the proposed grading system, has higher accuracy in comparison with the models with patients classified by Watari et al (9) grading system for all the GRF feature combinations. The highest accuracy (94.78%) was achieved by the ECM classifier trained with GRF features extracted from the GRFx and GRFy components for the proposed grading model. Also, it is observed that the proposed grading model outperformed the Watari grading system for ML-based DSPN severity classifications for all possible combinations of GRF features during gait. So, from this comparative analysis, it can be concluded that our proposed grading model is a suitable real-life application for DSPN patient's severity identification using clinical data (MNSI) and also using biomechanical data during gait (EMG and GRF).

IV. DISCUSSION

As per our knowledge, this is the first ML-based work to classify patients with different degrees of DSPN severity using biomechanical data during gait. In this study, we have collected biomechanical data (EMG and GRF) from the Watari et al (9) study, where we have a total of 77 patients' data from four experimental classes, (absent, mild, moderate and severe) and observed the EMG and GRF characteristics during gait. The patients were graded into different severity classes by the MNSI grading system proposed in their study. After the raw data pre-processing, the feature extraction technique, previously proposed by Johirul et al (23) was used, where two new time-domain features (LMAV, and NSV) with other 17 existing time-domain features were combined. This new feature extraction scheme showed better performance compared with other reported feature extraction schemes in the literature. For both EMG and GRF, correlation among the extracted features was calculated using pairwise linear correlation and if the correlation coefficient is greater than equals to 0.9, the among two features the second features were eliminated from the feature list. After removing highly correlated features, the Relieff feature ranking technique was used to rank the features based on their impact on identifying the different level of DSPN severity. After finalizing the features for ML model training, we find out the best-performing model.

Table VII
PERFORMANCE ANALYSIS OF ML MODEL USING DIFFERENT COMBINATIONS OF FEATURES EXTRACTED FROM THREE GRF COMPONENTS DURING GAIT FOR PROPOSED GRADING SYSTEM BY WATARI ET AL. (9).

| Channel | GRF Components considered | Features | Accuracy (%) | Sensitivity (%) | Specificity (%) | Precision (%) | F1-Score (%) | AUC |
|---|---|---|---|---|---|---|---|---|
| One Channel | GRFx | Top 12 | 76.76±3.87 | 76.77±3.80 | 90.84±2.93 | 76.77±3.80 | 76.86±3.33 | 0.88 |
| | GRFy | Top 9 | 80.11±2.27 | 80.12±2.30 | 93.43±2.27 | 80.12±2.33 | 80.66±2.60 | 0.92 |
| | GRFz | Top 11 | 82.22±2.94 | 82.25±2.91 | 92.95±2.21 | 82.25±2.91 | 82.46±3.09 | 0.92 |
| Two Channel | GRFx, GRFz | Top 15 | 85.39±2.53 | 85.40±2.46 | 94.59±3.49 | 85.40±2.46 | 85.91±2.19 | 0.94 |
| | GRFx, GRFy | Top 15 | 86.62±3.26 | 86.58±3.23 | 95.29±2.35 | 86.58±3.23 | 86.71±3.35 | 0.95 |
| | GRFy, GRFz | Top 15 | 85.21±1.89 | 85.24±1.85 | 94.83±2.94 | 85.24±1.85 | 85.59±1.86 | 0.95 |
| Three Channel | GRFx, GRFy, GRFz | Top 18 | 86.62±2.75 | 86.59±2.78 | 95.07±3.5 | 86.59±2.78 | 86.95±2.89 | 0.95 |

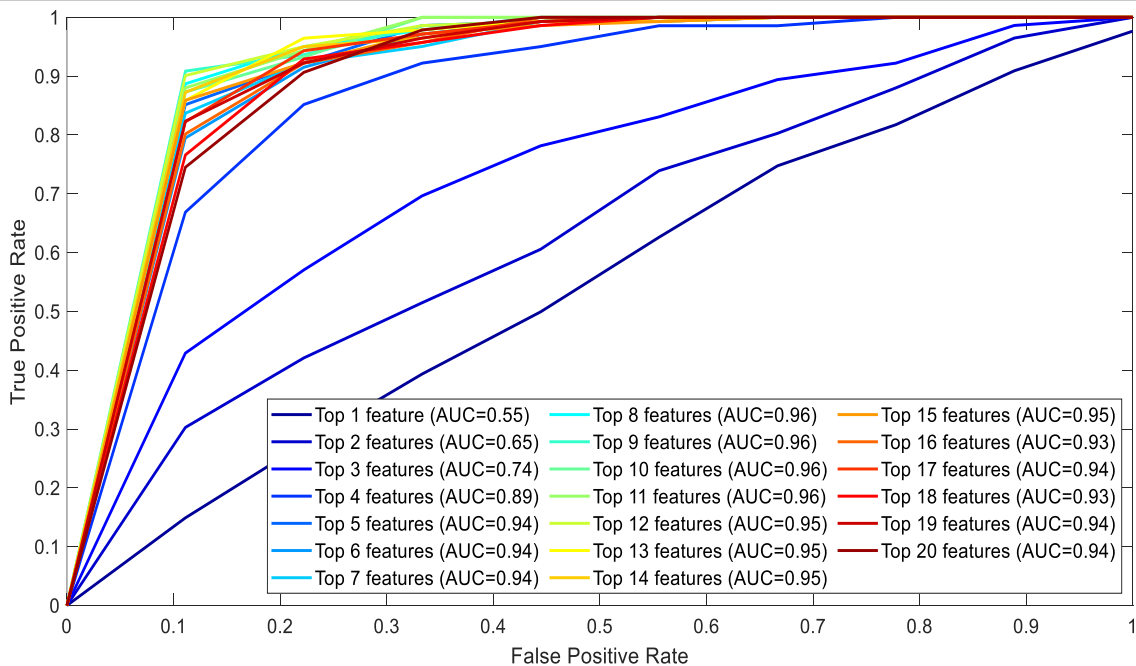

Fig. 11. ROC curve for top-ranked feature combinations from GRFx and GRFy signal for the proposed grading system by Watari et al. (9)

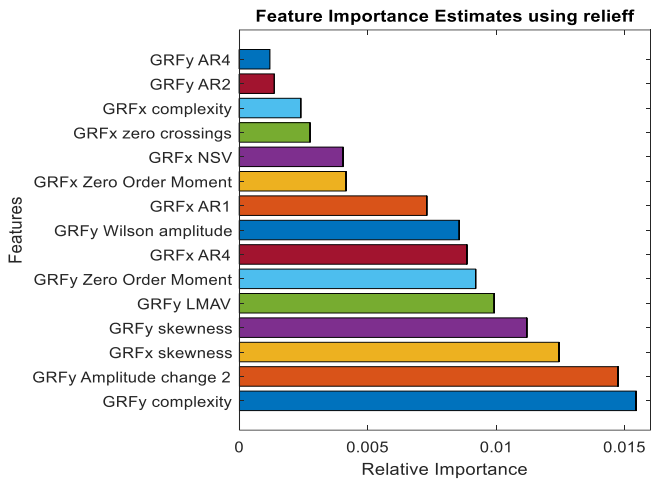

Fig. 12. Top-ranked features extracted combined from GRFx, and GRFy, components signal by Relieff feature selection techniques for the proposed grading system (22).

We have trained 6 different conventional machine learning algorithms using both EMG, and GRF features to find out the best performing algorithm. Among these 6 ML models, ECM exhibited the best performance with both EMG and GRF features. After finding the best performing algorithm, we tuned the hyperparameters of the ECM model and trained the model for different combinations of features and their performance was evaluated. In ML learning, model optimization has a great influence on the supervised learning technique. As the ML model learned from the data, the tuned model helped in better learning and enhancing the performance of the classification models.

This study has been conducted for three sub-studies for EMG, and GRF signals. In One channel analysis, we have considered the features from only one lower limb muscle or GRF component, in the two-channel, we considered the feature combinations of two muscles or GRF components, and in the three-channel, all the muscles or GRF component features were considered. Depending on the channel, the number of features was varied which feed to the ML model for training. In this study, we have trained our ML model in all possible ways to find out the best-performing features.

TABLE VIII
PERFORMANCE ANALYSIS OF ML MODEL USING DIFFERENT COMBINATIONS OF GRF FEATURES FROM THREE GRF COMPONENTS DURING GAIT FOR THE PROPOSED (22) GRADING SYSTEM.

| Channel | GRF Components considered | Features | Accuracy (%) | Sensitivity (%) | Specificity22 (%) | Precision (%) | F1-Score (%) | AUC |
|---|---|---|---|---|---|---|---|---|
| One Channel | GRFx | Top 13 | 90.11±1.73 | 90.11±1.73 | 97.48±0.66 | 90.11±1.73 | 90.20±1.6 | 0.95 |
|  | GRFy | Top 10 | 89.11±1.5 | 89.11±1.50 | 95.40±1.84 | 89.11±1.5 | 89.20±1.44 | 0.966 |
|  | GRFz | Top 10 | 86.44±1.27 | 86.44±1.27 | 95.11±2.49 | 86.44±1.28 | 86.58±1.26 | 0.97 |
| Two Channel | GRFx, GRFz | Top 15 | 94.78±2.31 | 94.78±2.31 | 97.93±2.06 | 94.78±2.31 | 94.858±2.28 | 0.99 |
|  | GRFx, GRFy | Top 16 | 92.56±3.05 | 92.56±3.05 | 97.33±2.89 | 92.56±3.05 | 92.68±3.02 | 0.98 |
|  | GRFy, GRFz | Top 13 | 93.78±1.85 | 93.78±1.85 | 96.74±1.70 | 93.78±1.85 | 93.86±1.81 | 0.98 |
| Three Channel | GRFx, GRFy, GRFz | Top 15 | 94.67±2.10 | 94.67±2.1 | 97.33±0.84 | 94.67±2.10 | 94.69±2.12 | 0.98 |

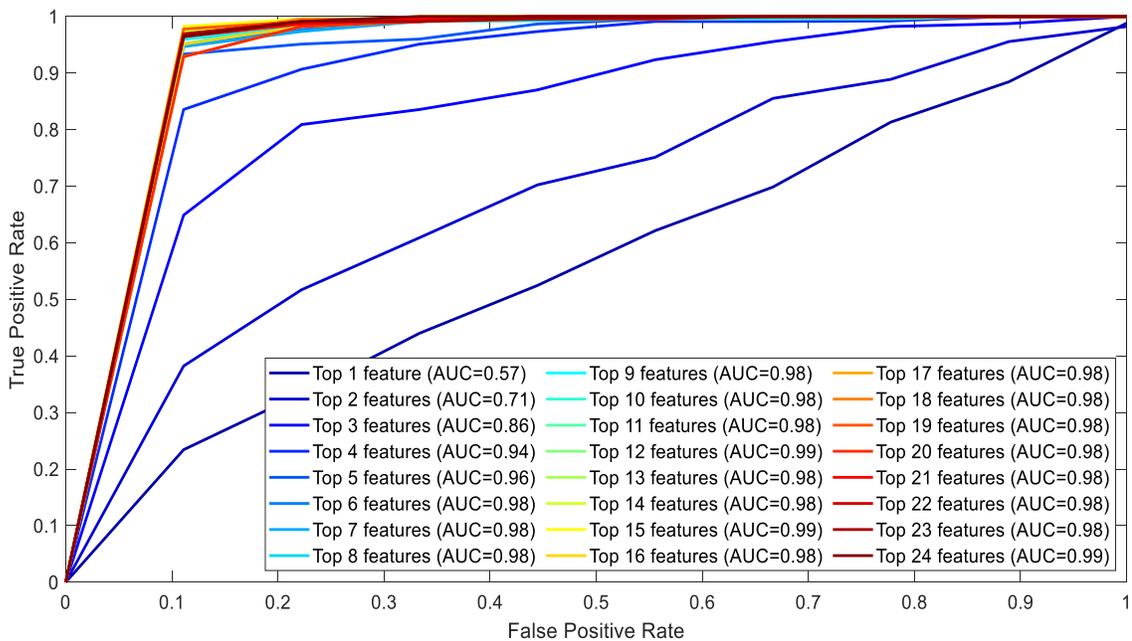

Fig. 13. ROC curve for top-ranked feature combinations from GRFx, GRFy and GRFz signals for the proposed grading system (22).

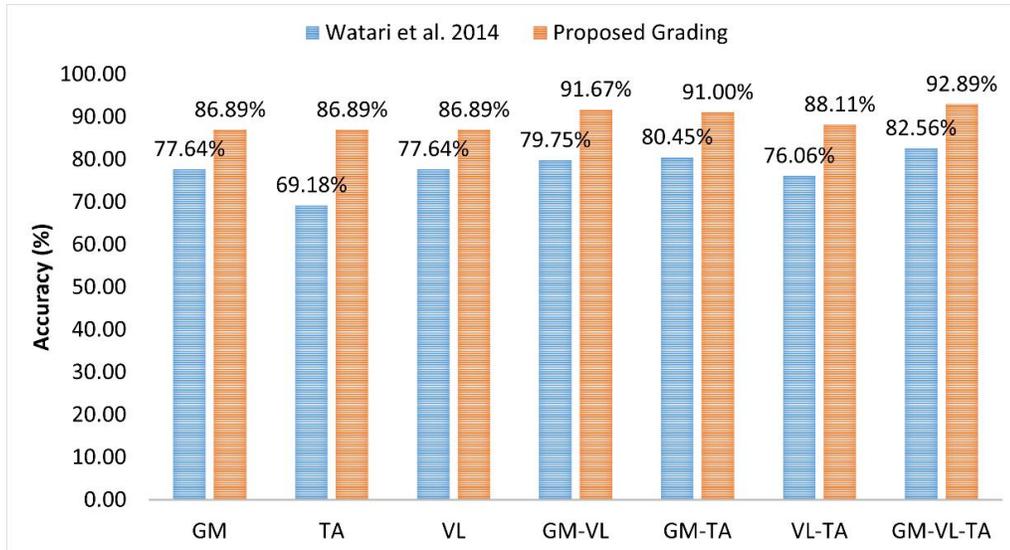

Fig. 14. Comparison between two grading systems [9, 22] based on the accuracy of the ML model trained using EMG features from three different lower limb muscles in different combinations.

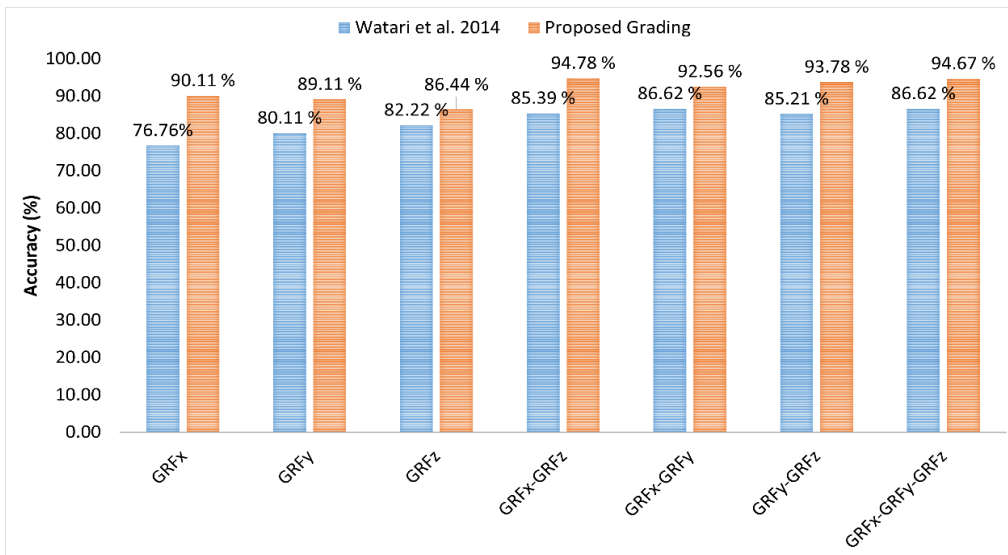

Fig. 15. Comparison between two grading systems [9, 22] based on the accuracy of the ML model trained using 3-dimensional GRF components features during gait in different combinations.

This study has been dealt with as a four-class problem, where we have considered four severity classes Absent, Mild, Moderate and Severe. The evaluation matrixes were calculated depending on the accurate identification of the right class. For EMG analysis, we have found the best accuracy of 82.56% using the Top 13 features from the three-channel analysis of GM, VL, and TA. In the GRF analysis, the model showed 86.61% accuracy by using the Top 15 from the combination of the features of the GRFx-GRFy signal. In this study, before training out ML models, we have pre-processed the dataset, in a very detailed way involving removing noise, and proper filtration, normalization, and segmentation of the signals, which made this data set without any biasing or overfitting issues. However, even though a thorough analysis has been conducted for ML-based DSPN severity classification using biomechanical features, the highest performance was achieved at only 86.61% by the features from GRFx and GRFy combined. This indicates that ML models were unable to distinguish between different severity classes with higher accuracy, which indicates that the patient's severity by Watari et al (9) is not accurately correlated with the biomechanical features. In other words, the grading is misleading ML model training. Therefore, the grading system proposed by Watari et al (9) cannot be considered for the reliable identification of the DSPN severity of the patients.

Initially, this work observes the performance of different ML-based DSPN severity classification models using biomechanical (EMG and GRF) features, where patients were graded using the grading system from the literature. Different data pre-processing, feature extraction, ranking, and diminution reduction methods were studied before developing the ML models. Different ML algorithms were trained, and their performance was analyzed to find out the best performing model for DSPN severity classification. Later, the performance of the proposed grading system on different ML-based classification problems was studied for biomechanical data. A comparative analysis was performed between the proposed grading method and the grading method from literature for MNSI.

Using the proposed grading model to classify patients' DSPN severity, all the analysis previously done with the existing grading system (9) was repeated for both EMG and GRF features, to validate the performance of the proposed grading system for ML-based DSPN severity classification. The best performing models using biomechanical features from EMG have an accuracy of 92.89% with the top 14 features, and for GRF features, the model achieved an accuracy of 94.78% with the Top 15 features combination, extracted from GRFx, GRFy and GRFz. The performance of ML-based DSPN severity classification models, improved significantly, indicating their reliability in DSPN severity classification, for biomechanical data. For over 50 years, analyses are being done to observe the biomechanical changes in gait due to DSPN. However, no literature was found to use ML using the biomechanical data to identify DSPN severity. The only literature was found where they used K-mean clustering to identify DSPN and Non-DSPN and control patients, however, it reported poor performance. This is the first work where DSPN severity was classified using the biomechanical data and with the elaborate pre-processing, ML models exhibited satisfactory performance, which is another novel contribution toward reliable DSPN severity classification using ML.

With the help of the proposed grading technique, three ML-based DSPN severity, classification models were recommended. For the EMG data-based DSPN severity classification model, the Top 14 features combination, extracted from the three lower limb muscles, ranked by relief feature ranking techniques, trained with ECM algorithm, along with the proposed grading system can be a complete solution. Similarly, using the Top 15 feature combinations from GRFx and GRFy components, ranked by relief technique, by relief feature ranking techniques, trained with ECM algorithm, along with the proposed grading system can be another complete solution. In this study, before training out ECM models, the dataset was preprocessed, in a very detailed way involving removing noise, and proper filtration, normalization, and segmentation of the signals, which made this data set without any biasing or overfitting issues. With the proposed grading system (22), ML models showed promising and reliable performance for DSPN identification for EMG and GRF features.

There is ample scope to study ML DSPN severity classification using biomechanical data. This is one of the very fast works applying ML in diagnosing and severity stratification of DSPN. In this study, 19-time domain features were used to extract valuable information from the EMG and GRF signals. In future, other feature selection, reduction and extraction techniques should be studied in identifying DSPN. Also, in this research, signal processing has been done manually. In future, automatic signal processing is needed to be incorporated for real-time detection. Performance analysis of different deep learning models should also be studied for biomechanical data for DSPN severity classification. Finally, in conclusion, this work is the beginning of ML-based application in DSPN severity identification and stratification. There are many unrevealed scopes of work yet to be discovered.

**Declarations:**

**Ethics approval and consent to participate:** Not applicable.

**Consent for publication:** Not applicable.

**Availability of data and materials:** The dataset used in this work is not publicly available.

**Competing interests:** The authors declare that they have no competing interests.

**Funding:** This research is financially supported by Universiti Kebangsaan Malaysia (UKM), Grant Number DIP-2020-004, GUP-2021-019 and Qatar National Research Foundation (QNRF) grant no. NPRP12s-0227-190164 and International Research Collaboration Co-Fund (IRCC) grant: IRCC-2021-001 and ASEAN-India Collaborative Research Project, Department of Science and Technology, Science and Engineering Research Board (DST-SERB), Govt. of India, Grant No. CRD/2020/000220. The statements made herein are solely the responsibility of the authors.

**Authors' contributions:** Conceptualization: F. H. and M.E.H.C; Data curation: S.H.M.A., A.A.A.B., M.A.S.B. and M.H.H.M.; Formal analysis: F. H., M.A., E.H. and M.S.; Investigation: F. H., M.A., E.H. and M.S.; Methodology: F. H., M.A., E.H. and M.S.; Project administration: M.B.I.R., M.E.H.C, S.K. and G.S.; Resources: M.B.I.R., M.E.H.C, S.H.M.A. E.K. and A.A.A.B.; Supervision: M.B.I.R., M.E.H.C and S.K.; Validation: F. H., M.B.I.R., M.E.H.C, M.A.S.B. and M.H.H.M.; Visualization: S.H.M.A., A.A.A.B., G.S. and M.H.H.M.; Manuscript Writing: F. H.; Manuscript review & editing, All the authors reviewed the manuscript.

**Acknowledgements:** Authors would like to thank Isabel C.N. Sacco, (Physical Therapy, Speech, and Occupational Therapy Department, School of Medicine) and Ricky Watari, Universidade de São Paulo for providing us with the dataset from their study.